\documentclass[letterpaper, 10 pt, conference]{ieeeconf}  

\usepackage{amssymb}
\usepackage{pifont}
\newcommand{\cmark}{\ding{51}}%
\newcommand{\xmark}{\ding{55}}%

\usepackage{amsmath,bm}
\usepackage[utf8]{inputenc} 
\usepackage[T1]{fontenc}    
\usepackage{hyperref}       
\usepackage{url}            
\usepackage{booktabs}       
\usepackage{amsfonts}       
\usepackage{nicefrac}       
\usepackage{microtype}      
\usepackage{subcaption}
\usepackage{caption}
\usepackage{graphicx}
\usepackage[ruled,vlined]{algorithm2e}
\usepackage{MnSymbol}

\usepackage[bibencoding=utf8,sorting=none,maxnames=50,firstinits=false]{biblatex}
\IEEEoverridecommandlockouts                              

\title{\LARGE \bf A Reinforcement Learning Approach for Robotic Unloading from Visual Observations}

\author{Vittorio Giammarino*$^{1}$, Alberto Giammarino$^{2}$ and Matthew Pearce$^{3}$
\thanks{*Part of this work was done while Giammarino was an intern in the AI team at Pickle Robot Co., 1280 Cambridge St., Cambridge, MA 02139,USA}
\thanks{$^{1}$Vittorio Giammarino is with the Division of Systems Engineering, Boston University, Boston, MA 02446, USA, {\tt\small vgiammar@bu.edu}}%
\thanks{$^{2}$Alberto Giammarino is an independent researcher {\tt\small alberto.giammarino96@gmail.com}}
\thanks{$^{3}$Matthew Pearce is in the AI team at Pickle Robot Co., 1280 Cambridge St., Cambridge, MA 02139, USA, {\tt\small \{matt\}@picklerobot.com}}}

\begin{document}

\maketitle
\thispagestyle{empty}
\pagestyle{empty}

\begin{abstract}
In this work, we focus on a robotic unloading problem from visual observations, where robots are required to autonomously unload stacks of parcels using RGB-D images as their primary input source. While supervised and imitation learning have accomplished good results in these types of tasks, they heavily rely on labeled data, which are challenging to obtain in realistic scenarios. Our study aims to develop a \emph{sample efficient} controller framework that can learn unloading tasks \emph{without the need for labeled data} during the learning process. To tackle this challenge, we propose a hierarchical controller structure that combines a high-level decision-making module with classical motion control. The high-level module is trained using Deep Reinforcement Learning (DRL), wherein we incorporate a safety bias mechanism and design a reward function tailored to this task. Our experiments demonstrate that both these elements play a crucial role in achieving improved learning performance. Furthermore, to ensure reproducibility and establish a benchmark for future research, we provide free access to our \href{https://github.com/VittorioGiammarino/RL-for-unloading-from-pixels}{code and simulation}.
\end{abstract}

\section{Introduction}

Robotic unloading is generally defined as the set of tasks in which robots are deployed to unload items from containers, trucks, or other transportation vehicles. The successful progress of this technology represents a compelling opportunity for the future, as it can address various challenges encountered in logistics, manufacturing, and warehousing. Within these industries, unloading tasks involve physically demanding and repetitive actions that can pose risks for human workers. In this regard, robotic unloading offers a way to enhance workers' safety and mitigate hazards associated with heavy lifting and challenging work environments. 

\begin{figure*}[h!]
    \centering
    \begin{subfigure}[t]{0.49\textwidth}
        \centering
        \includegraphics[width=4cm, height=2cm]{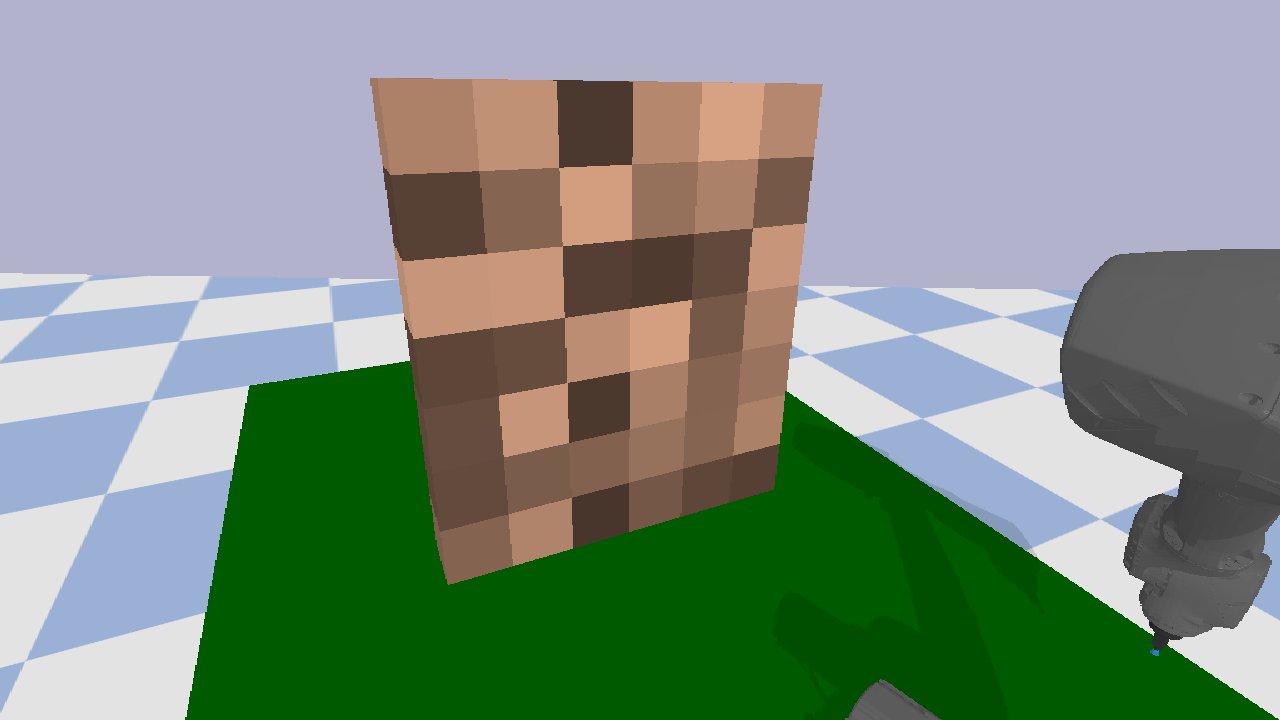}
        \caption{Visual Observation.}
        \label{fig:obs}
    \end{subfigure}
    ~
    \begin{subfigure}[t]{0.49\textwidth}
        \centering
        \includegraphics[width=4cm, height=2cm]{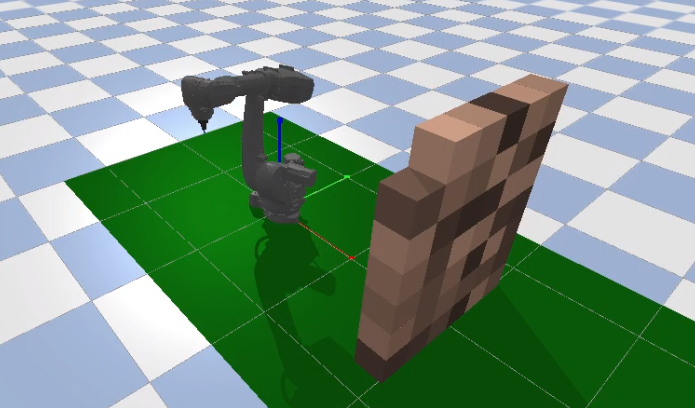}
        \caption{Environment.}
        \label{fig:task}
    \end{subfigure}
    \caption{Overview of our robotic unloading setup. Fig.~\ref{fig:obs} illustrates a snapshot of the visual observations available to the robot for the decision-making process. These observations are captured by a camera positioned on the top-left side of the robot and consist in $720 \times 1280$ high resolution RGB-D images. Fig.~\ref{fig:task} presents an overview of the unloading task. Stacks of parcels are positioned in front of an industrial KUKA KR70 robotic manipulator equipped with a suction gripper. Using the visual information depicted in Fig.~\ref{fig:obs}, the robot has to select a parcel to pick, grasp it, and then place it on the ground on its right side. The parcels in the scene are randomized in terms of color within the brown spectrum. A video demonstrating the task is available in the Supplementary Materials.}
    \label{fig:task+obs}
\end{figure*}

In this paper, we investigate robotic unloading tasks from visual observations (see Fig.~\ref{fig:task+obs} for an overview of the environment). Specifically, our objective is to enable a robotic manipulator to autonomously unload stacks of parcels by using RGB-D images as primary input source. We formulate this problem as a three-dimensional pick-and-place task, where the parcels, arranged in piles, are picked from the stack and placed on a floor conveyor. Previous studies have addressed pick-and-place by integrating objects' pose estimation \cite{yoon2003real, zhu2014single} with scripted planning and motion control \cite{frazzoli2005maneuver}. While these systems demonstrate robustness in structured environments, they are unsuitable for uncertain and unstructured settings, which require improved generalization capabilities. In order to address these challenges, recent years have witnessed a surge of interest in machine learning techniques. In particular, end-to-end Reinforcement Learning (RL) has been successfully used for pick-and-place in \cite{levine2016end, kalashnikov2018scalable}. However, end-to-end RL faces difficulties in real-world scenarios due to the large amount of data required to achieve acceptable performance. To improve data efficiency, recent work has integrated RL with object-centric assumptions such as keypoints \cite{manuelli2019kpam, kulkarni2019unsupervised}, embeddings \cite{jund2018optimization} or dense descriptors \cite{florence2018dense, florence2019self}. These representations are typically learned through supervised learning \cite{liu2011supervised}, which often involves tedious and expensive data labeling or annotation processes. Another line of research has explored Imitation Learning (IL), also known as learning from demonstrations \cite{zeng2020transporter, huang2022equivariant, shridhar2022cliport}. Despite achieving promising results, IL remains a supervised learning approach that relies on collecting expert demonstrations, which is akin to collecting labeled data and can be costly, time-consuming, or even infeasible. As a result, the main goal of this paper is to develop a \emph{sample efficient} controller framework that can learn the robotic unloading task in Fig.~\ref{fig:task+obs}, \emph{without requiring any form of labeled data}.    

Towards addressing this problem, we propose a hierarchical controller structure that separates the robot decision-making process from the low-level module. The decision-making component, named high-level controller, is trained using DRL~\cite{sutton2018reinforcement, franccois2018introduction} and more specifically Deep Q-Learning (DQL) \cite{mnih2013playing} from RGB images. Meanwhile, the low-level module relies on classical trajectory planning and motion control techniques. Within this framework, our work makes two main contributions. 

From an algorithmic perspective, our main novelties lie in the high-level controller, aiming to improve the sample efficiency of our DQL pipeline. First, we equip our Deep Q-Networks with a \emph{safety bias mechanism} with the goal of biasing the decision policy towards safe end-effector configurations. Note that an end-effector configuration is considered safe if it is reachable, i.e., it is contained within the robot workspace. Additionally, we propose a task-specific reward function \emph{which takes into account the verticality} of our unloading task. In order to test the impact of these mechanisms on learning performance, we conduct an ablation study and show how both these elements are crucial to accomplish improved results in our task. 

Our second contribution involves the development of a simulated environment using the PyBullet physics simulator \cite{coumans2016pybullet}. Simulators play a crucial role in prototyping RL algorithms as they offer a cost-effective and risk-free testing environment. This aspect is particularly valuable for industry-oriented tasks like robotic unloading, which are challenging to replicate in a research setting. Having a simulator available becomes essential in facilitating new research in this domain. Therefore, we provide open access to our environment, with the goal of establishing an interesting benchmark for future studies in this field.  

The remainder of the paper is organized as follows: Section~\ref{sec:related_work} provides a summary of the work related to robotic unloading. Section~\ref{sec:preliminaries} introduces notation and background on RL. Section~\ref{sec:simulated_env} provides a detailed description of the simulation environment. Section~\ref{sec:methods} presents the hierarchical controller and outlines the algorithm used to train the high-level controller. Finally, Section~\ref{sec:results} presents our experimental results and Section~\ref{sec:conclusion} concludes the paper providing a general discussion on our findings.    

\section{Related Work}
\label{sec:related_work}
In the following, we briefly review the most relevant work on robotic unloading.
One of the earliest studies in this field is presented in \cite{scholz2008development}, which offers an overview of the main technical challenges associated with automatic unloading. This study proposes solutions based on a 3D laser scanner perception system and scripted trajectories. 

In recent years, there has been a significant focus on leveraging deep learning (DL) techniques to enhance perception systems tailored to robotic unloading tasks. Papers such as \cite{hwang2022object, medrano2022box} formulate algorithms for accurate object detection and parcel segmentation within the context of robotic unloading. However, these studies primarily concentrate on perception and do not address the decision-making and control aspects of the robot.

Other studies have explored the integration of parcel segmentation with robotic control problems. In \cite{stoyanov2016no} for instance, perception, planning, and control are individually addressed and subsequently combined within a unified framework. In particular, a RGB-D camera is attached to the robot gripper for perception, and segmentation models are utilized to identify the safest object in the scene. A customized motion planning framework is then employed for control. Similarly, in \cite{doliotis20163d}, a 3D vision processing algorithm is introduced for parcel segmentation, enabling on-the-fly determination of multiple gripping strategies. 

Regarding the decision-making problem, papers such as \cite{islam2020planning, das2021learning} introduce a reasoning framework based on decision trees for optimal unloading, which is then combined with motion planning. However, these papers do not explicitly address the perception problem. Similarly, in \cite{park2021reinforcement}, a Q-learning-based algorithm is proposed for optimal decision-making, assuming accurate detection, perfect stacking, and uniform-sized boxes.

Compared to these studies, our research takes a comprehensive approach to the unloading problem. We explore controllers that integrate perception, decision-making, and motion control as an interconnected system and treat this interconnection as a unified problem.

\section{Preliminaries}
\label{sec:preliminaries}
Unless indicated otherwise, we use uppercase letters (e.g., $S_t$) for random variables, lowercase letters (e.g., $s_t$) for values of random variables, script letters (e.g., $\mathcal{S}$) for sets and reference frames, bold lowercase letters (e.g., $\bm{\theta}$) for vectors and bold uppercase letters (e.g. $\bm{H}$) for matrices. We denote expectation as $\mathbb{E}[\cdot]$.

Our decision-making process is modeled as a finite-horizon discounted Markov Decision Process (MDP) described by the tuple $(\mathcal{S}, \mathcal{A}, \mathcal{T}, \mathcal{R}, \rho_0, \gamma)$, where $\mathcal{S}$ is the set of states and $\mathcal{A}$ is the set of actions. $\mathcal{T}:\mathcal{S}\times \mathcal{A} \rightarrow P(\mathcal{S})$ is the transition probability function where $P(\mathcal{S})$ denotes the space of probability distributions over $\mathcal{S}$, $\mathcal{R}:\mathcal{S}\times \mathcal{A} \rightarrow \mathbb{R}$ is the reward function which maps state-action pairs to scalar rewards, $\rho_0 \in P(\mathcal{S})$ is the initial state distribution, and $\gamma \in [0,1)$ the discount factor. We define the agent as a stationary policy $\pi:\mathcal{S}\rightarrow P(\mathcal{A})$, where $\pi(a|s)$ is the probability of taking action $a$ in state $s$. When a function is parameterized with parameters $\bm{\theta} \in \varTheta \subset \mathbb{R}^k$ we write $\pi_{\bm{\theta}}$.

\paragraph{Reinforcement learning} Given an MDP and a stationary policy $\pi:\mathcal{S} \to P(\mathcal{A})$, the RL objective is to learn an optimal policy, $\pi^*$, which maximizes the expected total discounted reward 
\begin{align}
\begin{split}
    J(\pi) = \mathbb{E}_{\tau}\Big[\sum_{t=0}^{T}\gamma^t \mathcal{R}(s_t,a_t)\Big],
    \label{eq:RL_obj}
\end{split}
\end{align}
where $\tau = (s_0,a_0,s_1,a_1,\dots, a_{T-1},s_T)$ are trajectories sampled according to $s_0 \sim \rho_0$, $a_t\sim\pi(\cdot|s_t)$ and $s_{t+1}\sim \mathcal{T}(\cdot|s_t,a_t)$, and $T$ is the number of decision steps in a single episode. Note that, neither the transition probability function $\mathcal{T}$ nor the reward function $\mathcal{R}$ are available to the learning agent. Therefore, the agent interacts with the environment, collects transitions $(s_t, a_t, r_t, s_{t+1})$, where $r_t=\mathcal{R}(s_t,a_t)$, and exploits these transitions to estimate $J(\pi)$ in \eqref{eq:RL_obj} in order to learn $\pi^*$ as $\pi^* \in \arg\max_{\pi} J(\pi)$. 

Well-known algorithms solve the RL problem by estimating, and optimizing, the value functions induced by the policy $\pi$. We define the state value function as $V^{\pi}(s)= \mathbb{E}_{\tau}[\sum_{t=0}^{T}\gamma^t \mathcal{R}(s_t,a_t)|S_0=s]$ and the state-action value function as $Q^{\pi}(s,a) = \mathbb{E}_{\tau}[\sum_{t=0}^{T}\gamma^t \mathcal{R}(s_t,a_t)|S_0=s, A_0=a]$. Furthermore, $V^{\pi}(s) = \mathbb{E}_{a\sim\pi(\cdot|s)}[Q^{\pi}(s,a)]$. Note that, the optimal policy $\pi^* \in \arg\max_{\pi} J(\pi)$ induces the optimal state value function
\begin{equation*}
    V^*(s) = \max_{\pi} V^{\pi}(s), \ \ \ \forall s \in \mathcal{S}.
\end{equation*}
Therefore, given $V^*(s)$ and assuming $\mathcal{T}$ is known, we can retrieve $\pi^*(s)$ for all $s \in \mathcal{S}$ as the action $a\in\mathcal{A}$ that leads to the highest expected return $\mathbb{E}_{s_{t+1}\sim\mathcal{T}(\cdot|s_t, a)}[V^*(s_{t+1})]$ for all $s_t \in \mathcal{S}$. 

In the RL setting, where $\mathcal{T}$ is unknown, Temporal Difference (TD) methods \cite{sutton2018reinforcement} compute $V^*(s)$ and $\pi^*(s)$ by leveraging the following iterative updating rule for $Q^{\pi}(s_t, a_t)$:
\begin{align}
    Q^{\pi}(s_t, a_t) \leftarrow Q^{\pi}(s_t, a_t) + \alpha(Y - Q^{\pi}(s_t, a_t)),
    \label{eq:TD_methods}
\end{align}
where $\alpha$ is a learning rate and $Y$ is a target value as in a standard regression problem. 

Given a set of transitions $(s_t, a_t, r_t, s_{t+1}, a_{t+1})$, generated by the interactions of the policy $\pi$ with the environment; setting $Y = r_t + \gamma Q^{\pi}(s_{t+1}, a_{t+1})$ yields the on-policy update typical of SARSA \cite{sutton2018reinforcement}. On the other hand, setting $Y = r_t + \gamma \max_{a}Q^{\pi}(s_{t+1}, a)$ yields the off-policy update used in $Q$-learning \cite{sutton2018reinforcement}. Provided the optimal state-action value function $Q^{*}(s,a)$, then $V^*(s) = \max_a Q^*(s,a)$ and $\pi^*(s) \in \arg\max_{a} Q^*(s,a)$ for all $s \in \mathcal{S}$.

\section{Simulated Environment}
\label{sec:simulated_env}
\begin{figure*}
    \centering
    \includegraphics[width=\linewidth, height=2.5cm]{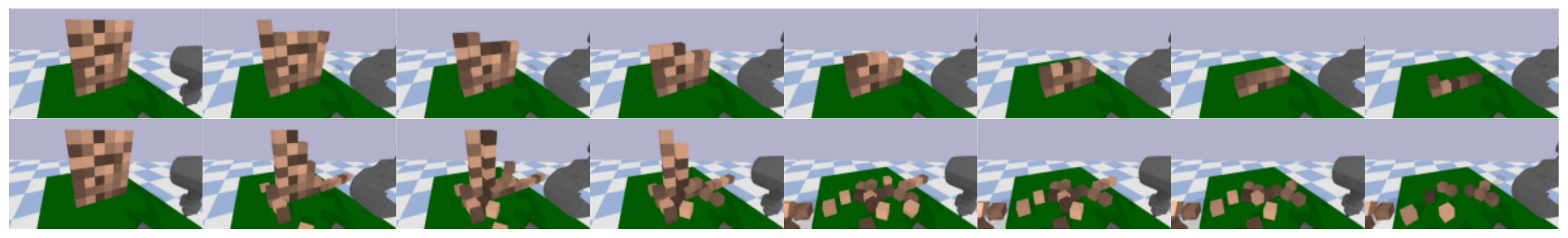}
    \caption{The evolution of the scene is illustrated in two different episodes. In the upper row episode, the agent follows an optimal decision policy, successfully unloading all the parcels. In the lower row episode, the policy is suboptimal and a single wrong decision impacts the subsequent organization of the scene, affecting the agent's final outcome and undermining the entire unloading process.}
    \label{fig:scene}
\end{figure*}

In the following we provide a detailed description of our robotic unloading task. The environment is summarized in Fig.~\ref{fig:task+obs} and is based on the PyBullet physics simulator \cite{coumans2016pybullet}. The simulated environment can be accessed at our \href{https://github.com/VittorioGiammarino/RL-for-unloading-from-pixels}{GitHub repository}\footnote{https://github.com/VittorioGiammarino/RL-for-unloading-from-pixels}. 

\paragraph{Agent} The decision agent controls an industrial $6$-axis KUKA KR70 robotic manipulator. The robot inertial frame corresponds to the world frame and is denoted with $\mathcal{I}$. We refer to the end-effector pose with the homogeneous transformation $\boldsymbol{H}_{\mathcal{I}\mathcal{E}}$, where $\mathcal{E}$ is the robot end-effector frame. The robot end-effector is a suction gripper with a contact sensor which enables the contact detection between the gripper and the other objects of the scene. The unloading task relies on the visual observations illustrated in Figure~\ref{fig:obs} and \ref{fig:scene}. These images are captured by a RGB-D camera positioned at the top-left side of the robot, providing $720\times1280$ high-resolution images. The intrinsic and extrinsic parameters of the camera are known, and along with the depth image, are used to transform from the 2D pixel space to the robot inertial frame $\mathcal{I}$ \cite{szeliski2010computer}. Additionally, the agent is provided with the robot workspace boundaries in $\mathcal{I}$, denoted with $W_{\mathcal{I}}$. 

\paragraph{Task} For each episode, our task involves unloading a stack of $42$ parcels, arranged as shown in Fig.~\ref{fig:task+obs}. Each parcel is a cube with $0.25m$ edge length and randomized color surface within the brown spectrum. At each decision step, the agent is required to select a parcel, grasp it, and then place it on the ground, where a floor conveyor will complete the unloading. The complexity of this problem, compared to more classical pick-and-place or rearrangement tasks, arises from the sequential nature of the decision-making process and the limited margin of error allowed by the task. In this regard, Fig.~\ref{fig:scene} illustrates two episodes generated by different decision policies. In the upper row, a successful unloading sequence is depicted, where the agent successfully unloads all the $42$ parcels. Conversely, the lower row shows an unsuccessful unloading sequence, where a single decision has a negative impact on the subsequent scene, ultimately compromising the agent's final outcome. Note that the simulation environment is modelled as a finite-horizon MDP with $T=42$, i.e., the number of decision steps is equivalent to the number of initialized parcels at $t=0$. When a mistake is made, as shown in Fig.~\ref{fig:scene} (lower), we remove the $n$ parcels that land outside the agent's workspace $W_{\mathcal{I}}$, and the agent incurs a time penalty of $n$, i.e. $t \leftarrow t+n$ rather than $t \leftarrow t+1$. Similarly, when the agent attempts picks out of $W_{\mathcal{I}}$, a parcel is removed from the top of the stack and $t \leftarrow t+1$. In other words, the agent is forced to skip a step. This ensures $T=42$ for all the episodes, preventing infinite loops during the training process. Note that infinite loops, like repeatedly choosing parcels outside of $W_{\mathcal{I}}$, tend to occur frequently in the initial stages of the learning process and can prevent progress if not addressed.

\section{Methods}
\label{sec:methods}
\begin{figure}
    \centering
    \includegraphics[width=0.45\textwidth]{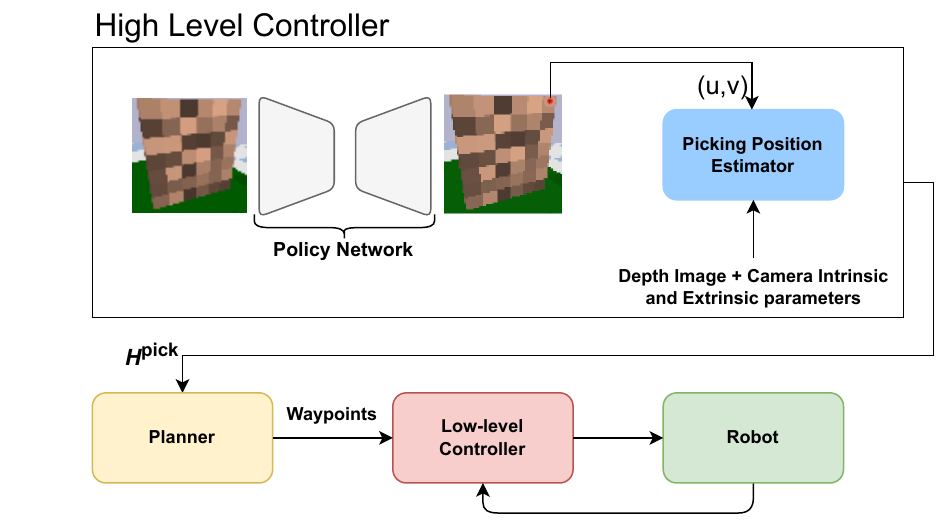}
    \caption{Summary of our hierarchical controller. The high-level controller selects a pixel $a = (u,v)$ from an observation and computes a picking pose denoted with $\bm{H}^{\text{pick}}_{\mathcal{I}\mathcal{E}}(a)$. This pose is passed to a planner, which generates a trajectory of Cartesian waypoints. The low-level controller transforms this Cartesian trajectory into a joint space trajectory solving an inverse kinematics problem. The resulting trajectory serves as a reference for the PD controller of the actuators.}
    \label{fig:controller}
\end{figure}

In this section, we introduce our hierarchical controller framework, consisting of a high-level decision-making module and a low-level trajectory planning and motion control module (cf. Fig.~\ref{fig:controller}). Moreover, we provide a comprehensive description of our DQL pipeline, used to train the agent's decision-making policy. 

\paragraph{High-level controller} Given a visual observation $s_t \in \mathcal{S}$, the goal of the high-level controller is to provide a corresponding picking end-effector pose, denoted by the homogeneous transformation $\bm{H}^{\text{pick}}_{\mathcal{I}\mathcal{E}}$, which maximizes the expected number of unloaded parcels over the episode. Hence, at each decision step $t$, we want
\begin{equation}
    f(s_t)  \to \bm{H}^{\text{pick}}_{\mathcal{I}\mathcal{E}}, \ \ \ t \in [0,T).
    \label{eq:high_level_controller}
\end{equation}

The function $f$ in \eqref{eq:high_level_controller} is defined as a composition of two main steps. In the first step, a policy $\pi_{\bm{\theta}}: \mathcal{S} \to P(\mathcal{A})$ takes a $64\times64$ RGB image of the scene and selects a single pixel $a_t = (u,v)_t$. The $64\times64$ image is a cropped and resized version of the original $720\times1280$ image in Fig.~\ref{fig:obs}. Note that the original image is cropped assuming prior knowledge of the area containing the stack of parcels during the unloading process. Using an aligned depth image, the selected $a_t = (u,v)_t$ is transformed into Cartesian coordinates, representing the picking end-effector position in $\mathcal{I}$ denoted with $\bm{p}^{\text{pick}}_{\mathcal{I}\mathcal{E}}(a_t)$. The picking position is then associated with a rotation matrix $\bm{C}^{\text{pick}}_{\mathcal{I}\mathcal{E}}(a_t)$, corresponding to the picking end-effector orientation. Picking position $\bm{p}^{\text{pick}}_{\mathcal{I}\mathcal{E}}(a_t)$ and orientation $\bm{C}^{\text{pick}}_{\mathcal{I}\mathcal{E}}(a_t)$ yield the picking end-effector pose $\bm{H}^{\text{pick}}_{\mathcal{I}\mathcal{E}}(a_t)$, where, compared to the notation in \eqref{eq:high_level_controller}, we explicitly state the dependency on $a_t=(u,v)_t$. In this work, we assume a fixed end-effector orientation $\bm{C}^{\text{pick}}_{\mathcal{I}\mathcal{E}}(a_t)$, which is computed to be orthogonal to the plane of the stack front surface (the visible surface in Fig.~\ref{fig:obs}, \ref{fig:scene} and~\ref{fig:controller}). It is important to note that $\bm{p}^{\text{pick}}_{\mathcal{I}\mathcal{E}}(a_t)$ and $\bm{C}^{\text{pick}}_{\mathcal{I}\mathcal{E}}(a_t)$ are interdependent choices. In our setting, we condition $\bm{p}^{\text{pick}}_{\mathcal{I}\mathcal{E}}(a_t)$ on $\bm{C}^{\text{pick}}_{\mathcal{I}\mathcal{E}}(a_t)$, i.e., $\bm{p}^{\text{pick}}_{\mathcal{I}\mathcal{E}}(a_t)$ is learned given $\bm{C}^{\text{pick}}_{\mathcal{I}\mathcal{E}}(a_t)$. The alternative setting, where $\bm{C}^{\text{pick}}_{\mathcal{I}\mathcal{E}}(a_t)$ is conditioned on $\bm{p}^{\text{pick}}_{\mathcal{I}\mathcal{E}}(a_t)$, is left for future work.

\paragraph{Learning algorithm} The picking policy $\pi_{\bm{\theta}}$ is learned via DQL \cite{mnih2013playing}. We define $K$ critic networks as $Q_{\bm{\theta}_k}: \mathcal{S} \times \mathcal{A} \to \mathbb{R}$, parameterized by a $43$-layer encoder-decoder residual network (ResNet) \cite{he2016deep} where the same encoder is shared across all the critic networks.

The critic networks output picking confidence values over the pixel space and are trained to minimize:
\begin{align}
    \mathcal{L}_{\bm{\theta}_k}(\mathcal{B}) &= \mathbb{E}_{(s_t,a_t,r_t,s_{t+1})\sim\mathcal{B}}[(y_t - Q_{\bm{\theta}_k}(s_{t},a_t))^2], \label{eq:MSE} \\
    y_t &= r_t + \gamma \max_{a}\min_{k}Q_{\bm{\bar{\theta}}_k}(s_{t+1},a), \label{eq:value} 
\end{align}
where $\mathcal{B}$ is a replay buffer (cf. \cite{mnih2013playing}) and $\bm{\bar{\theta}}_k$ are the slow moving weights for the target critic networks \cite{van2016deep}. We use multiple critics and target networks to address the optimistic bias introduced by function approximation \cite{thrun1993issues, fujimoto2018addressing}. Moreover, we define two different reward functions:
\begin{align}
\begin{split}
    \mathcal{R}_b(s,a) &= \begin{cases}
    1, \ \text{if picking succeeds}, \\
    0, \ \text{otherwise},
    \end{cases} \\
    \mathcal{R}_v(s,a) &= \begin{cases}
    1+\lambda \cdot z^{\text{pick}}_{\mathcal{I}\mathcal{E}}(a), \ \text{if picking succeeds}, \\
    0, \ \text{otherwise},
    \end{cases}
\end{split} 
\label{eq:reward_func}
\end{align}
where $\lambda$ is a scalar hyperparameter, and $z^{\text{pick}}_{\mathcal{I}\mathcal{E}}(a)$ is the $z$-coordinate in $\bm{p}^{\text{pick}}_{\mathcal{I}\mathcal{E}}(a)$ where the $z$-axis represents the up direction in $\mathcal{I}$. Both the functions in \eqref{eq:reward_func} provide rewards when picking succeeds. Moreover, $\mathcal{R}_v(s,a)$ adds a bonus proportional to the $z$-coordinate, i.e., the height, of the picking position. This bonus incentivizes the unloading of parcels when in the upper part of the stack. In other words, it disincentivizes behaviors which can lead to the collapse of the stack as in the lower row episode in Fig.~\ref{fig:scene}. We experimentally show that $\mathcal{R}_v(s,a)$ remarkably improves performance when compared to $\mathcal{R}_b(s,a)$ in our task.

The loss in \eqref{eq:MSE}-\eqref{eq:value} is computed by sampling interactions $(s_t, a_t, r_t ,s_{t+1})$ from a replay buffer $\mathcal{B}$. During training, the agent interacts with the environment according to the exploration policy summarized in Algorithm~\ref{alg:exploration_policy}. The safety bias layer, named Mask in Algorithm~\ref{alg:exploration_policy}, is defined as 
\begin{align}
    \text{Mask}(Q_{\bm{\bar{\theta}}_k}(s,a)) = \begin{cases}
    Q_{\bm{\bar{\theta}}_k}(s,a) + b, \ \ \ \text{if $\bm{p}^{\text{pick}}_{\mathcal{I}\mathcal{E}}(a) \in W_{\mathcal{I}}$}, \\
    Q_{\bm{\bar{\theta}}_k}(s,a), \ \ \ \text{otherwise},
    \end{cases}
    \label{eq:mask_func}
\end{align}
where $W_{\mathcal{I}}$ is the robot workspace and $b$ is a positive definite safety bias treated as an hyperparameter. The main goal of Mask in \eqref{eq:mask_func} is to bias the policy towards the actions with a corresponding $\bm{p}^{\text{pick}}_{\mathcal{I}\mathcal{E}}(a) \in W_{\mathcal{I}}$. We experimentally show that this layer becomes crucial to improve performance and efficiency in our task. During evaluation, given $s$, the actions are generated following $a \in \arg\max_a \text{Mask}(\min_k Q_{\bm{\bar{\theta}}_k}(s,a))$ where we use Mask in \eqref{eq:mask_func} as additional safety measure. We provide a summary of the full algorithm in Appendix.

\begin{algorithm}
\footnotesize
\caption{Exploration Policy}
\label{alg:exploration_policy}
\textbf{Input}: \\
$s$, $Q_{\bm{\bar{\theta}}_k}$, $\epsilon$, $\text{Mask}$, $W_{\mathcal{I}}$, $\mathcal{U}$, $\sigma$: state ($64 \times 64$ RGB image), target critic networks, exploration probability, safety bias layer, robot workspace, uniform distribution, and softmax function. \\
\Begin(ExplorationAction[$s$]){
$u \sim \mathcal{U}(0,1)$\\
\If{$u \leq \epsilon$}{
$a \sim \mathcal{U}\{a_{W_{\mathcal{I}}}\}$ where $a_{W_{\mathcal{I}}}$ denotes the action $a$ such that $\bm{p}^{\text{pick}}_{\mathcal{I}\mathcal{E}}(a) \in W_{\mathcal{I}}$
}
\Else{$a \sim \sigma(\text{Mask}(\min_k Q_{\bm{\bar{\theta}}_k}(s,a)))$}
\Return{$a$}
}
\end{algorithm}

\paragraph{Low-level module} The low-level module comprises a trajectory planner and a low-level controller. The planner receives $\bm{H}^{\text{pick}}_{\mathcal{I}\mathcal{E}}(a_t)$ from the high-level controller and provides a set of $5$ end-effector waypoints to the low-level controller. This set of waypoints consists of: $\bm{H}^{\text{pre-pick}}_{\mathcal{I}\mathcal{E}}$, $\bm{H}^{\text{pick}}_{\mathcal{I}\mathcal{E}}(a_t)$, $\bm{H}^{\text{post-pick}}_{\mathcal{I}\mathcal{E}}$, $\bm{H}^{\text{pre-place}}_{\mathcal{I}\mathcal{E}}$, $\bm{H}^{\text{place}}_{\mathcal{I}\mathcal{E}}$, $\bm{H}^{\text{out-of-camera}}_{\mathcal{I}\mathcal{E}}$. Specifically,  $\bm{H}^{\text{place}}_{\mathcal{I}\mathcal{E}}$ represents the placing pose which is assumed to be known throughout the task. $\bm{H}^{\text{out-of-camera}}_{\mathcal{I}\mathcal{E}}$ is the pose required for an unoccluded image of the scene, as shown in Fig.~\ref{fig:obs}. $\bm{H}^{\text{pre-pick}}_{\mathcal{I}\mathcal{E}}$ is the pre-picking pose, $\bm{H}^{\text{post-pick}}_{\mathcal{I}\mathcal{E}}$ the post-picking pose and $\bm{H}^{\text{pre-place}}_{\mathcal{I}\mathcal{E}}$ the pre-placing pose. Between $\bm{H}^{\text{pre-pick}}_{\mathcal{I}\mathcal{E}}$ and $\bm{H}^{\text{pick}}_{\mathcal{I}\mathcal{E}}(a_t)$, the gripper is activated as soon as a contact with a parcel is detected, after which the end-effector is moved towards $\bm{H}^{\text{pre-place}}_{\mathcal{I}\mathcal{E}}$.

Provided a trajectory of Cartesian waypoints, the low-level controller transforms this trajectory from Cartesian to joint space by solving an inverse kinematics optimization problem. The joint space trajectory is then used as a reference for the PD controller of the actuators.

\section{Experiments}
\label{sec:results}
\begin{figure}
    \centering
    \includegraphics[width=0.48\textwidth]{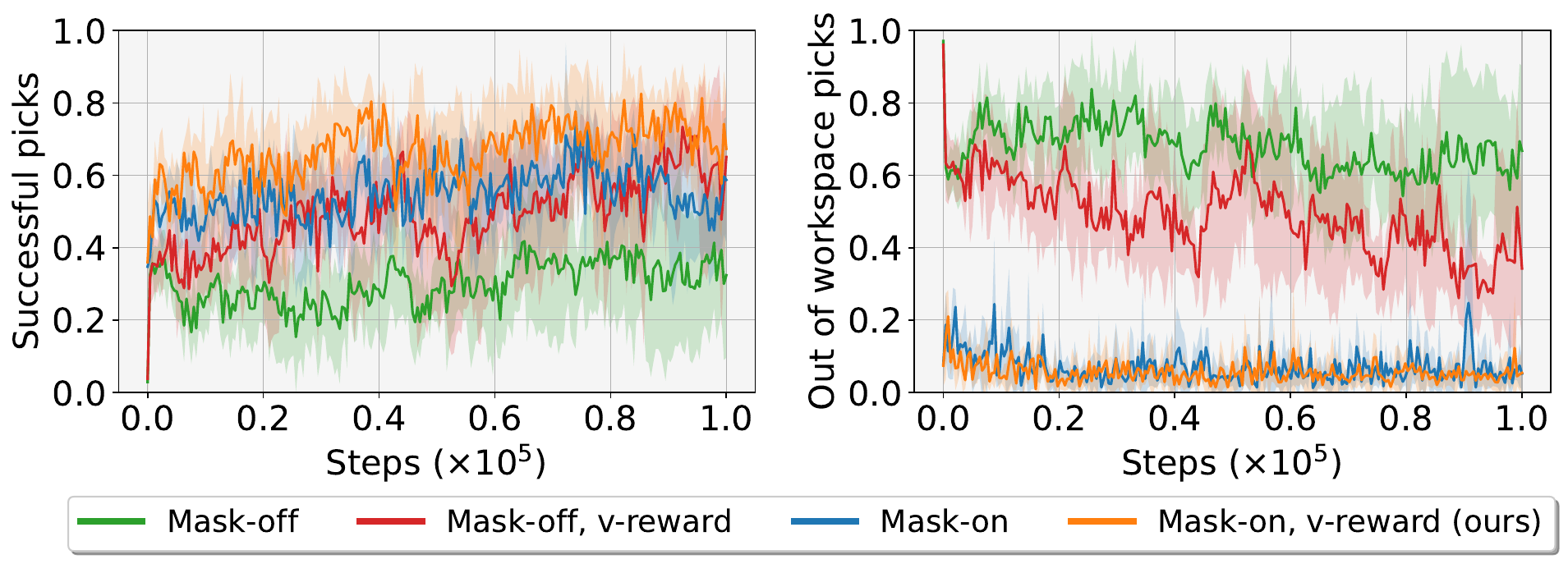}
    \caption{Ablation experiments. (Left) the average normalized number of successful picks, which represents the number of parcels successfully unloaded. (Right) the average normalized number of picks attempted with $\bm{p}^{\text{pick}}_{\mathcal{I}\mathcal{E}}(a) \notin W_{\mathcal{I}}$. Both results are averaged over $6$ seeds and the shaded area represents the standard deviation over seeds. For each seed, we randomly initialize the critic networks and train for $10^5$ steps. We evaluate the learned policy every $10$ training episodes, i.e., $420$ steps, using average performance over $3$ episodes. The characteristics of the tested algorithms are summarized in Table~\ref{table_1}.}
    \label{fig:experiment_average}
\end{figure}

All our experiments focus on the unloading problem described in Section~\ref{sec:simulated_env}. We evaluate four different versions of our DQL algorithm as summarized in Table~\ref{table_1}. It is important to note that both $\mathcal{R}_v$ in \eqref{eq:reward_func} and Mask in \eqref{eq:mask_func} contain two important hyperparameters, respectively $\lambda$ and $b$, which are kept fixed throughout our experiments. In particular, we set $\lambda=2$ and $b=100$. This choice for $\lambda$ takes into account the maximum height of the stack and ensures an appropriate trade-off between the two factors in $\mathcal{R}_v$. As for $b$, its optimal value depends on the initialization of the critic networks weights and on the considered reward function. The experiments depicted in Fig.~\ref{fig:experiment_average} show that $b=100$ yields good empirical results for all the algorithms in which Mask is used. 

\begin{table}
\centering
\scriptsize
\caption{A summary of the different algorithms tested in our experiments. Mask denotes the use of the safety bias layer in \eqref{eq:mask_func}, while $\mathcal{R}_v$ denotes the use of $\mathcal{R}_v$ rather than $\mathcal{R}_b$ in \eqref{eq:reward_func}.}
\label{table_1}
\begin{tabular}{l c c c c c c c c}\toprule
 & \multicolumn{2}{c}{\textit{Mask-off}} & \multicolumn{2}{c}{\textit{Mask-off, v-reward}} & \multicolumn{2}{c}{\textit{Mask-on}} & \multicolumn{2}{c}{\textit{Mask-on, v-reward}} \\
\cmidrule(lr){1-9}
Mask & \multicolumn{2}{c}{\xmark} & \multicolumn{2}{c}{\xmark} & \multicolumn{2}{c}{\cmark} & \multicolumn{2}{c}{\cmark} \\
$\mathcal{R}_v$ & \multicolumn{2}{c}{\xmark} & \multicolumn{2}{c}{\cmark} & \multicolumn{2}{c}{\xmark} & \multicolumn{2}{c}{\cmark} \\
\bottomrule
\end{tabular}
\end{table}

As commonly done in the RL literature, in our experiments we randomize training and evaluation episodes from the same distribution. The obtained final results are summarized in Fig.~\ref{fig:experiment_average} where the average normalized performance over $6$ random seeds is illustrated. Specifically, the left figure shows the average number of successful picks, and the right figure shows the number of attempted picks with $\bm{p}^{\text{pick}}_{\mathcal{I}\mathcal{E}}(a) \notin W_{\mathcal{I}}$. All the curves are normalized with respect to $42$, i.e., the maximum number of parcels per episode. These results demonstrate that both Mask and $\mathcal{R}_v$ remarkably improve performance, in particular when they are jointly used during training. Specifically, by using $\mathcal{R}_v$ rather than $\mathcal{R}_b$ in \eqref{eq:reward_func}, the agent receives a more accurate feedback about the characteristics of the task and requires fewer interactions to achieve better results. Furthermore, by using Mask, we are able to effectively reduce the number of attempted picks in which $\bm{p}^{\text{pick}}_{\mathcal{I}\mathcal{E}}(a)\notin W_{\mathcal{I}}$. This leads to an improved exploration strategy, as the agent mainly focuses on viable actions that lead to $\bm{p}^{\text{pick}}_{\mathcal{I}\mathcal{E}}(a) \in W_{\mathcal{I}}$. Conversely, when Mask is not used, the number of actions with $\bm{p}^{\text{pick}}_{\mathcal{I}\mathcal{E}}(a) \notin W_{\mathcal{I}}$ increases, leading to less effective exploration strategies and slower learning rate.

\begin{figure}
    \centering
    \includegraphics[width=0.47\textwidth]{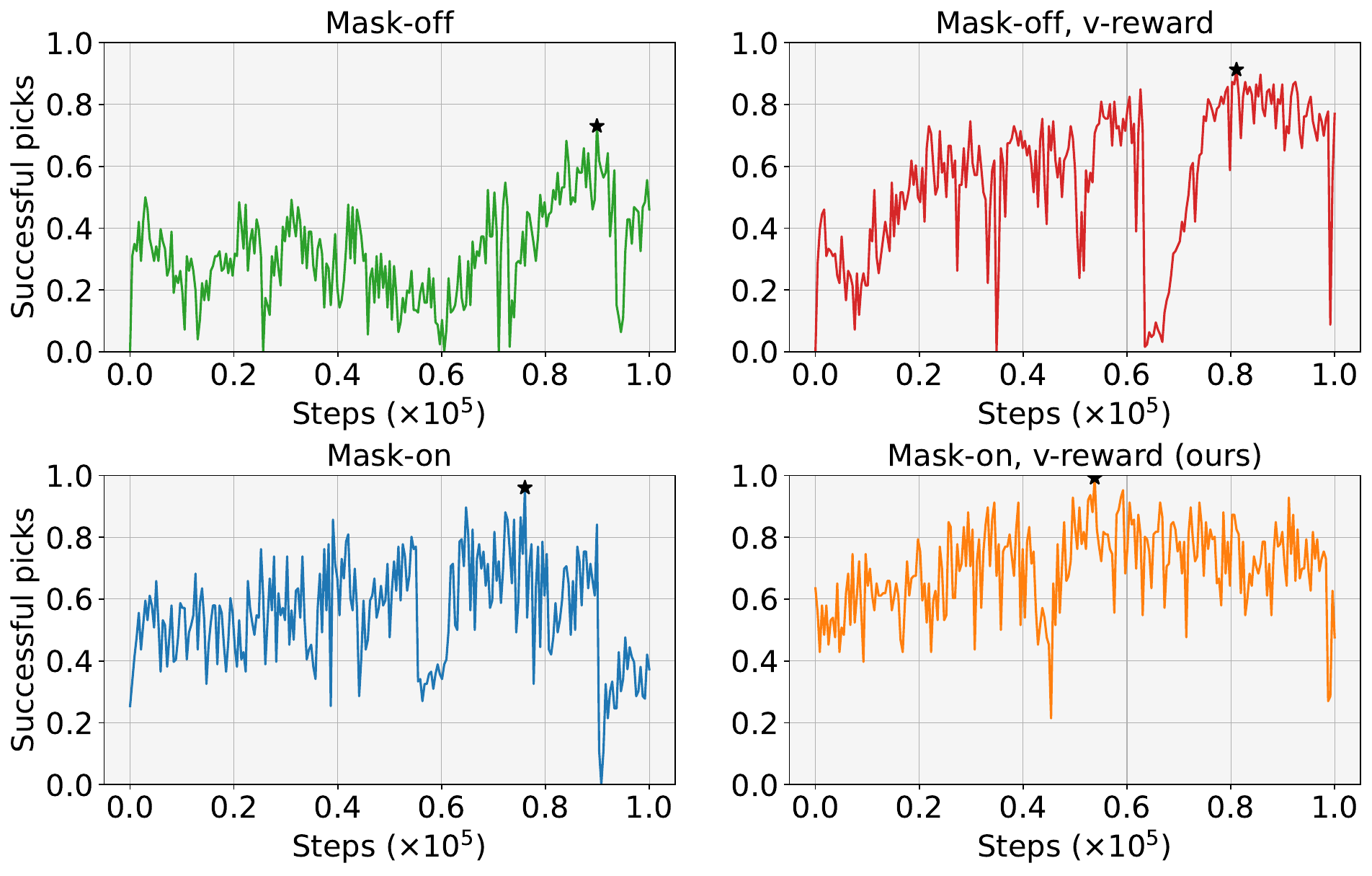}
    \caption{Training curves for the seeds with the best maximum performance. For each version of our algorithm, the best seed is chosen among the seeds averaged in Fig.~\ref{fig:experiment_average}. The maximum performance is highlighted with "$\filledstar$". The experiments are conducted as in Fig.~\ref{fig:experiment_average}. }
    \label{fig:experiment_best}
\end{figure}

In Fig.~\ref{fig:experiment_best}, we show, for each version of our algorithm, the seed leading to the best maximum performance among the seeds averaged in Fig.~\ref{fig:experiment_average}. These experiments more clearly emphasize the effect of Mask and $\mathcal{R}_v$ on improving our final results. In Fig~\ref{fig:experiment_best}, our best policy, trained with \textit{Mask-on, v-reward} (cf. Table~\ref{table_1}), achieves $99\%$ picking success over $3$ full evaluation episodes. 

The training curves in Fig.~\ref{fig:experiment_average} and Fig.~\ref{fig:experiment_best} are both summarized by the box plots in Fig.~\ref{fig:box_plot}. Additional results and all the used hyperparameters are provided in Appendix. We refer to our \href{https://github.com/VittorioGiammarino/RL-for-unloading-from-pixels}{GitHub repository} for more implementation details.

\begin{figure}
    \centering
    \begin{subfigure}[t]{0.4\textwidth}
        \centering
        \includegraphics[width=0.9\textwidth]{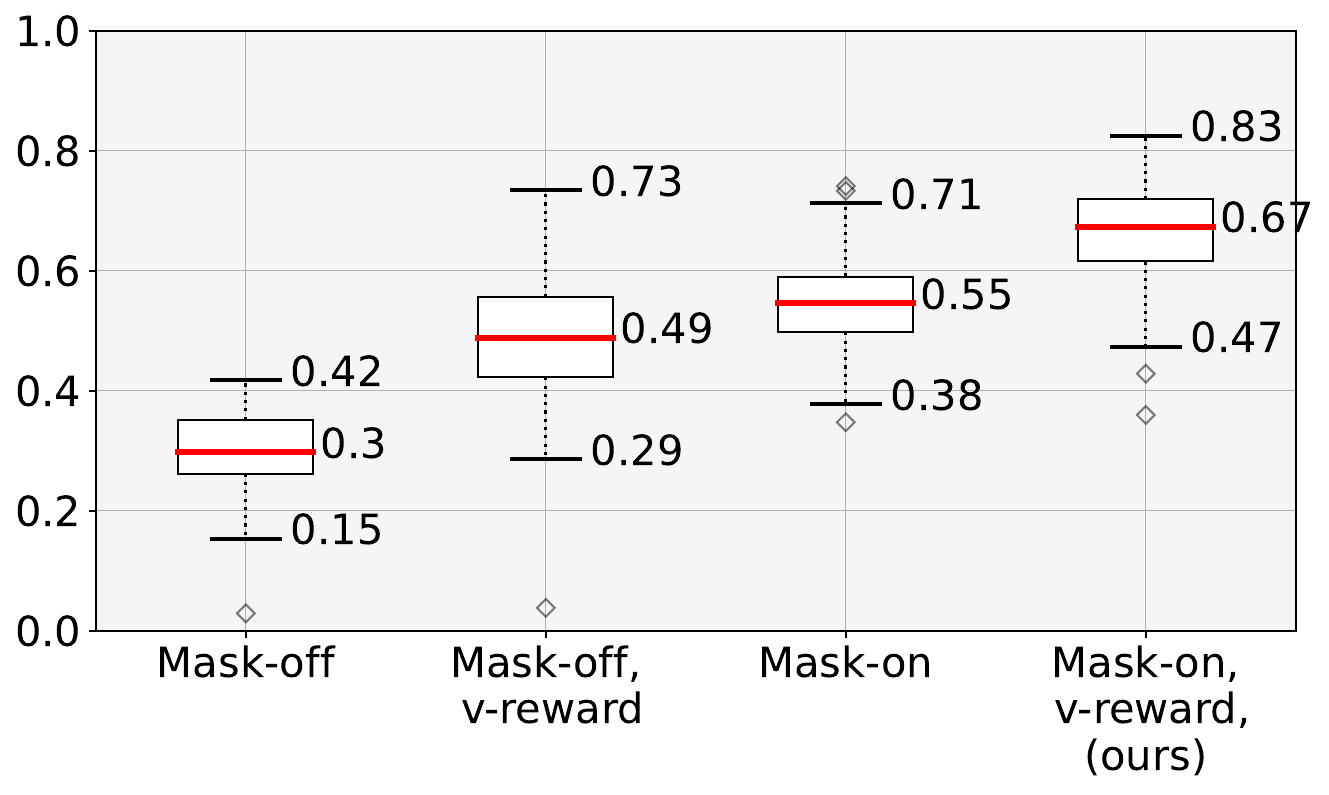}
        \caption{Box plot for the curves in Fig.~\ref{fig:experiment_average}.}
        \label{fig:box_plot_average}
    \end{subfigure}
    ~
    \begin{subfigure}[t]{0.4\textwidth}
        \centering
        \includegraphics[width=0.9\textwidth]{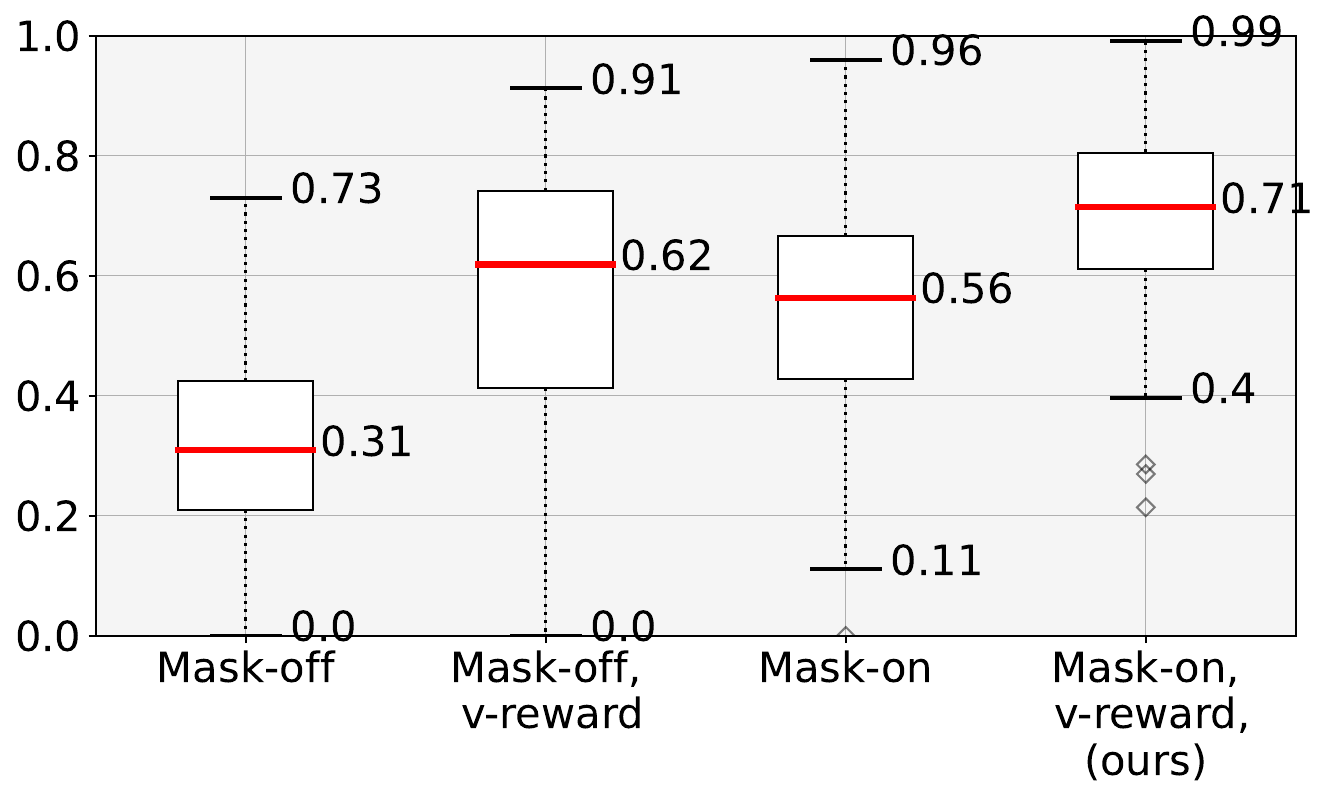}
        \caption{Box plot for the curves in Fig.~\ref{fig:experiment_best}.}
        \label{fig:box_plot_best}
    \end{subfigure}
    \caption{The box plots are generated considering the training curves in Fig.~\ref{fig:experiment_average} and Fig.~\ref{fig:experiment_best} where each evaluation step is treated as a data sample.}
    \label{fig:box_plot}
\end{figure}

\section{Conclusion}
\label{sec:conclusion}
In this study, we tackle the problem of robotic unloading from visual observations and propose a hierarchical controller that does not require any labeled data. Our hierarchical controller, depicted in Fig.~\ref{fig:controller}, consists of a high-level controller responsible for decision-making and a low-level module for trajectory planning and low-level control. Optimal decision-making is learned from RGB images using DQL and is converted in optimal end-effector pose leveraging an aligned depth image. In our DQL algorithm, we introduce a safety bias mechanism and a task-specific reward function which prove to be essential in order to improve our experimental results. Finally, we develop and make publicly available a simulated environment, with the goal of providing a benchmark for future studies in this field. 

\paragraph{Limitations and future work} Despite the advancement in addressing robotic unloading, it is important to understand the limitations of our approach. In the high level-controller, the main issue involves the training stability of our DQL algorithm. This problem is evident in Fig.~\ref{fig:experiment_best}, where the training progress does not show a monotonic trend. Furthermore, this instability is the main reason for the slow rate of improvement illustrated in Fig.~\ref{fig:experiment_average}. We consider this to be a crucial challenge for optimal unloading policy learning, representing a significant area for future research. 

Regarding the simulated environment, we consider it as a valuable benchmark for testing RL-based methods in unloading tasks. This perspective is substantiated by the results in Fig.~\ref{fig:experiment_average} and Fig.~\ref{fig:experiment_best}, where vanilla RL algorithms often struggle to succeed, as evidenced in the \textit{Mask-off} case. Moreover, prior research has addressed unloading tasks similar to what we present in our study \cite{islam2020planning, park2021reinforcement}. However, we acknowledge that our current simulation does not encompass all the potential variables encountered in real-world unloading scenarios. As a result, our future efforts will focus on improving our simulated environment in order to introduce more randomized settings, where the parcels can have different textures, size and shape. We emphasize this as an important research avenue towards improving performance in real-world scenarios, where parcel configurations are usually messy and uncertain. 

Looking ahead, we are also considering the prospect of directly applying our training solutions to real hardware in future research endeavors. This goal presents its own set of challenges, especially in tasks of this nature, where devising methods that ensure minimal human intervention and safety becomes of crucial importance.






\printbibliography
\clearpage

\section*{Appendix}

\section{Algorithm and hyperparameters}

The $K$ critic networks are defined as $Q_{\bm{\theta}_k}: \mathcal{S} \times \mathcal{A} \to \mathbb{R}$, and are parameterized by a $43$-layer encoder-decoder residual network (ResNet) \cite{he2016deep} with $12$ residual blocks. We use three $2$-stride convolutions in the encoder, which is shared among all the critic networks, and three bilinear-upsampling layers in each of the $K$ decoders. All the layers, except the last, are interleaved with ReLU activation functions. For more implementation details refer to our \href{https://anonymous.4open.science/r/RL-for-unloading-from-pixels-49F4}{GitHub repository}. Our DRL algorithm is summarized in Algorithm~\ref{alg:DQL_unloading} and the used hyperparameters are in Table~\ref{tab:Hyper_1}. 

\begin{algorithm}
\caption{Deep Q-Learning for unloading}
\label{alg:DQL_unloading}
\textbf{Input}: \\
$T_{\text{train}}$, $\zeta$, $B$, $\alpha$, $\gamma$, $K$, $Q_{\bm{\theta}_k}$, $\epsilon$, $\lambda$, $\text{Mask}$, $W_{\mathcal{I}}$, $\mathcal{U}$, $\sigma$: training steps, target update rate, batch size, learning rate, discount factor, number of critic networks, critic networks, exploration probability, $z$-coordinate picking weight, safety bias layer, robot workspace, uniform distribution, and softmax function. \\ 
\Begin(Training){
$s_0 \sim \rho_0$ \\
$\bar{\bm{\theta}}_k \leftarrow \bm{\theta}_k \ \ \ \forall k \in \{1,\dots,K\}$ \\
\For{$t=0, \dots, T_{\text{\normalfont{train}}}$}{
$a_t \leftarrow \text{ExplorationAction}[s_t]$ \\
$s_{t+1} \sim \mathcal{T}(\cdot|s_t, a_t)$ and $r_t = \mathcal{R}_v(s_{t}, a_t)$ \\
$\mathcal{B} \leftarrow \mathcal{B} \cup (s_t, a_t, r_t, s_{t+1})$ \\
UpdateCritic[$\mathcal{B}$] \\
$s_t \leftarrow s_{t+1}$ \\
}}
\Begin(ExplorationAction[$s$]){
$u \sim \mathcal{U}(0,1)$\\
\If{$u \leq \epsilon$}{
$a \sim \mathcal{U}\{a_{W_{\mathcal{I}}}\}$ where $a_{W_{\mathcal{I}}}$ denotes $a$ such that $\bm{p}^{\text{pick}}_{\mathcal{I}\mathcal{E}}(a) \in W_{\mathcal{I}}$
}
\Else{$a \sim \sigma(\text{Mask}(\min_k Q_{\bm{\bar{\theta}}_k}(s,a)))$}
\Return{$a$}
}
\Begin(UpdateCritic[$\mathcal{B}$]){
$\{(s_t, a_t, r_t, s_{t+1})\} \sim \mathcal{B}$ (sample $B$ transitions)\\ 
Update $Q_{\bm{\theta}_k}$ to minimize \eqref{eq:MSE}-\eqref{eq:value} with learning rate $\alpha$ \\
$\bar{\bm{\theta}}_k \leftarrow (1 - \zeta)\bar{\bm{\theta}}_k +\zeta\bm{\theta}_k \ \ \ \forall k \in \{1,\dots,K\}$
}
\end{algorithm}

\begin{table}
\centering
\caption{Hyperparameter values for our experiments.}
\label{tab:Hyper_1}
\small
\begin{tabular}{c c c c}\toprule
\multicolumn{2}{l}{Hyperparameter Name} & \multicolumn{2}{c}{Value}\\
\cmidrule(lr){1-2} \cmidrule(lr){3-4}
\multicolumn{2}{l}{Target update rate $(\zeta)$} & \multicolumn{2}{c}{$0.005$}\\
\multicolumn{2}{l}{Batch size $(B)$} & \multicolumn{2}{c}{$64$} \\
\multicolumn{2}{l}{Learning rate $(\alpha)$} & \multicolumn{2}{c}{$10^{-4}$}\\
\multicolumn{2}{l}{Discount factor $(\gamma)$} & \multicolumn{2}{c}{$0.99$} \\
\multicolumn{2}{l}{Number of critic networks $(K)$} & \multicolumn{2}{c}{$2$} \\
\multicolumn{2}{l}{Exploration probability $(\epsilon)$} & \multicolumn{2}{c}{$0.1$} \\
\multicolumn{2}{l}{$z$-coordinate picking weight ($(\lambda)$ in \eqref{eq:reward_func})} & \multicolumn{2}{c}{$2$}\\
\multicolumn{2}{l}{Safety bias ($(b)$ in \eqref{eq:mask_func})} & \multicolumn{2}{c}{$100$}\\
\multicolumn{2}{l}{Image size} & \multicolumn{2}{c}{$64 \times 64$} \\
\multicolumn{2}{l}{Optimizer} & \multicolumn{2}{c}{Adam}\\
\bottomrule
\end{tabular}
\end{table}

\newpage
\section{Additional Experiments}
\begin{figure*}
    \centering
    \begin{subfigure}[t]{0.8\linewidth}
        \centering
        \includegraphics[width=\textwidth]{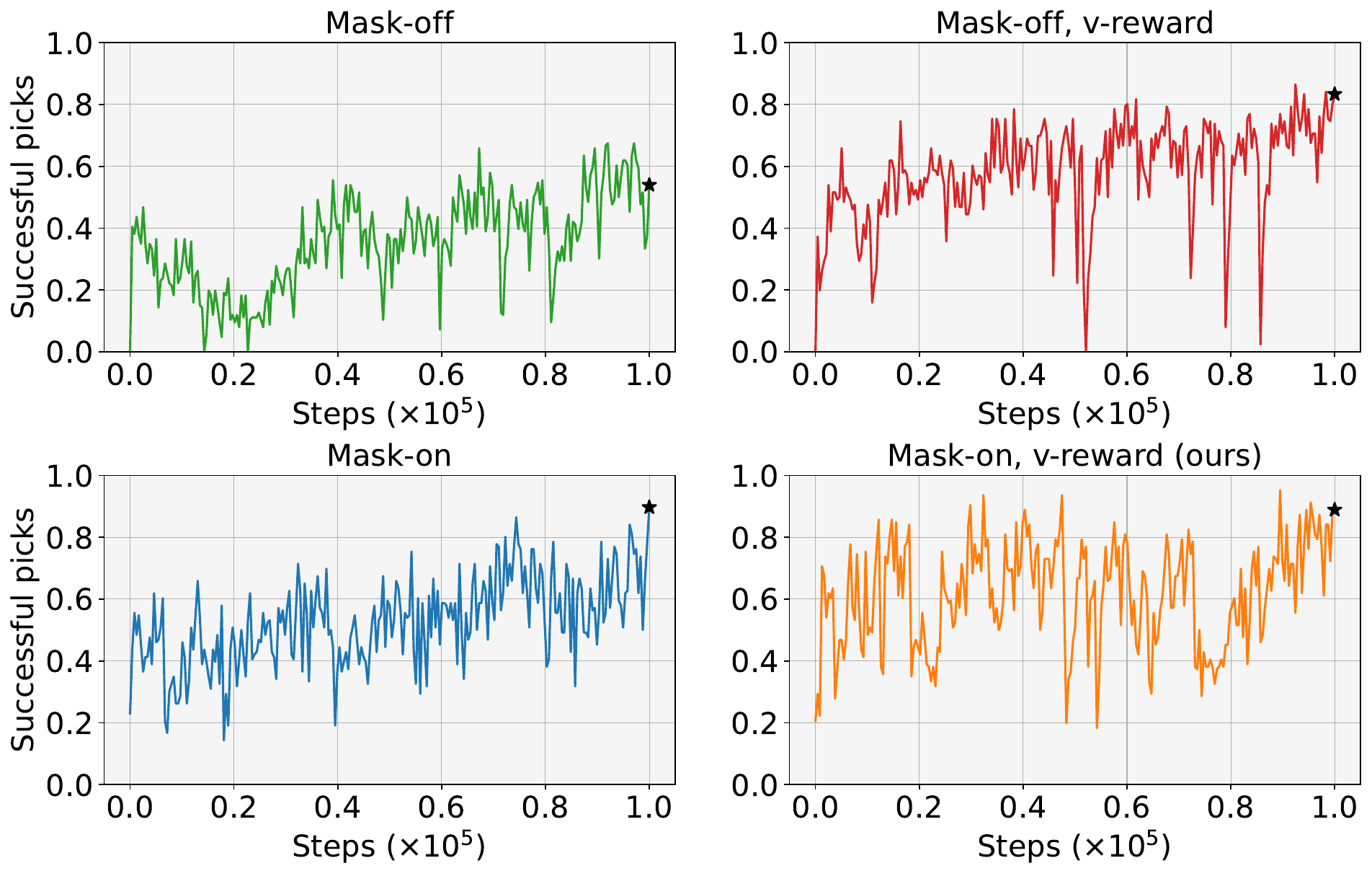}
        \caption{Training curves for the seeds with the best final performance.}
        \label{fig:best_final_app}
    \end{subfigure}
    ~
    \begin{subfigure}[t]{0.8\linewidth}
        \centering
        \includegraphics[width=0.7\textwidth]{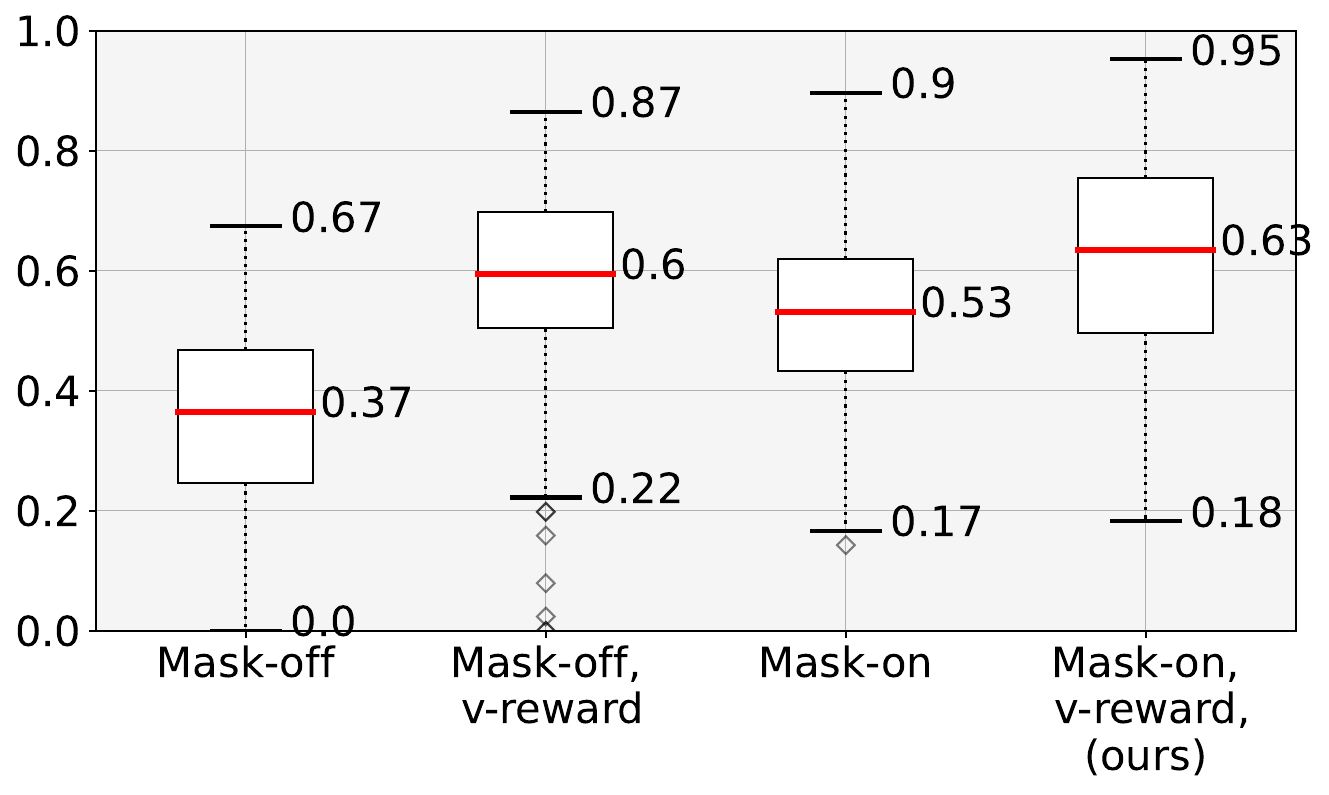}
        \caption{Box plot summarizing the results in Fig.~\ref{fig:best_final_app} where each evaluation step is treated as a data sample.}
        \label{fig:box_plot_best_final_app}
    \end{subfigure}
    \caption{Results for the seeds with the best final performance. For each version of our algorithm, the best seed is chosen among the seeds averaged in Fig.~\ref{fig:experiment_average}. The final performance is highlighted with "$\filledstar$". The experiments are conducted as in Fig.~\ref{fig:experiment_average} and the labels Mask-on and Mask-off denote the use of the safety bias layer in \eqref{eq:mask_func}, while v-reward denotes the use of $\mathcal{R}_v$ rather than $\mathcal{R}_b$ in \eqref{eq:reward_func}. }
    \label{fig:full_best_final results_app}
\end{figure*}

\begin{figure*}
    \centering
    \begin{subfigure}[t]{0.8\linewidth}
        \centering
        \includegraphics[width=\textwidth]{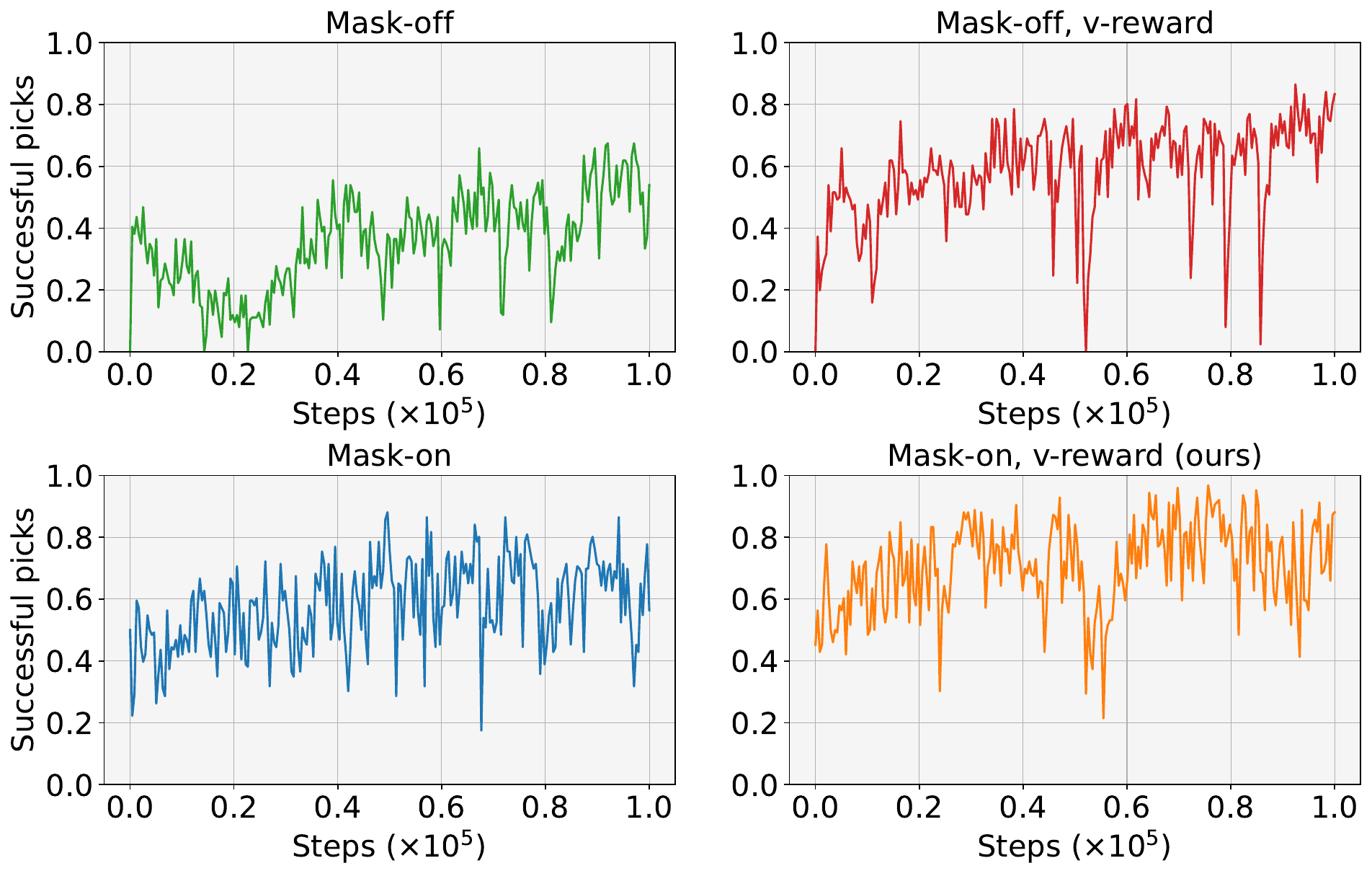}
        \caption{Training curves for the seeds with the best average performance during the training process.}
        \label{fig:best_mean_app}
    \end{subfigure}
    ~
    \begin{subfigure}[t]{0.8\linewidth}
        \centering
        \includegraphics[width=0.7\textwidth]{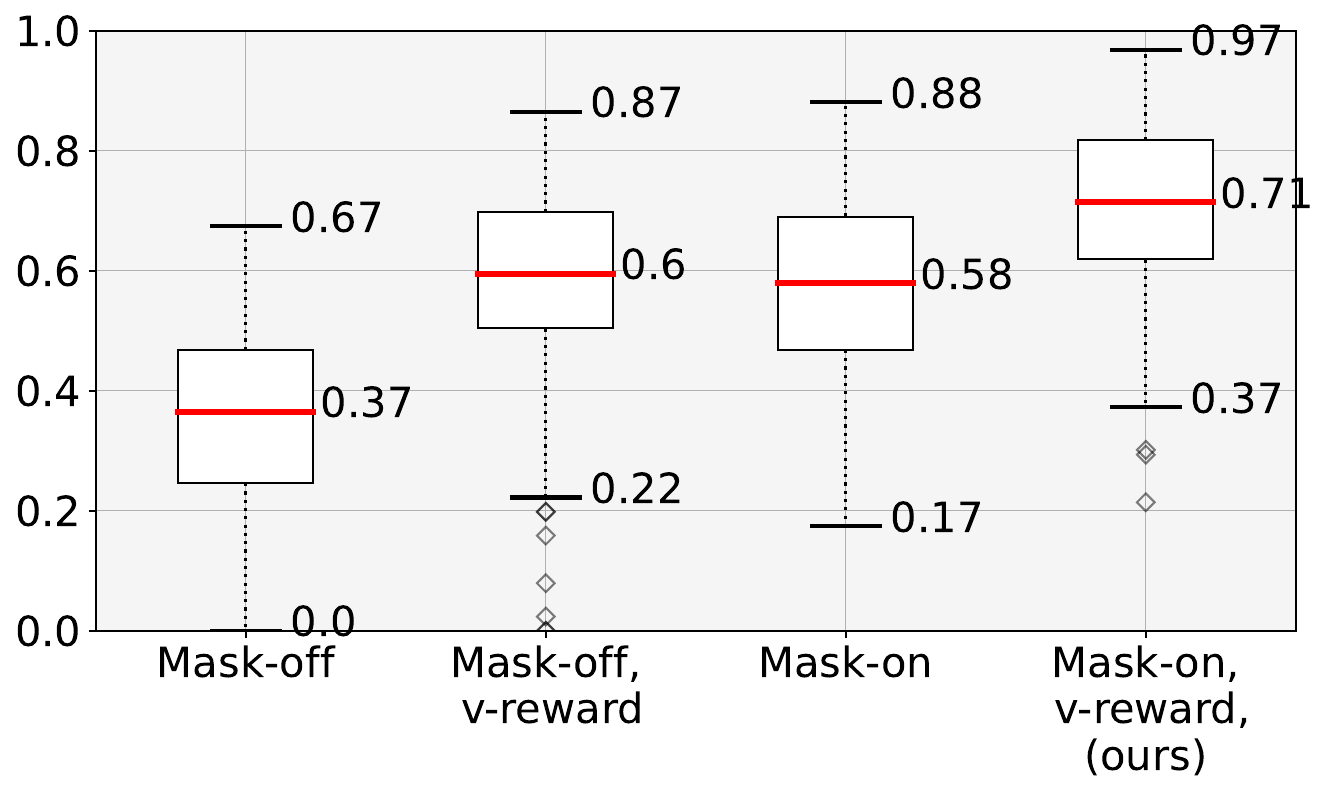}
        \caption{Box plot summarizing the results in Fig.~\ref{fig:best_mean_app} where each evaluation step is treated as a data sample.}
        \label{fig:box_plot_best_mean_app}
    \end{subfigure}
    \caption{Results for the seeds with the best average performance during the training process. For each version of our algorithm, the best seed is chosen among the seeds averaged in Fig.~\ref{fig:experiment_average}. The experiments are conducted as in Fig.~\ref{fig:experiment_average} and the labels Mask-on and Mask-off denote the use of the safety bias layer in \eqref{eq:mask_func}, while v-reward denotes the use of $\mathcal{R}_v$ rather than $\mathcal{R}_b$ in \eqref{eq:reward_func}.}
    \label{fig:full_best_mean_app}
\end{figure*}

\begin{figure*}
    \centering
    \begin{subfigure}[t]{0.8\linewidth}
        \centering
        \includegraphics[width=\textwidth]{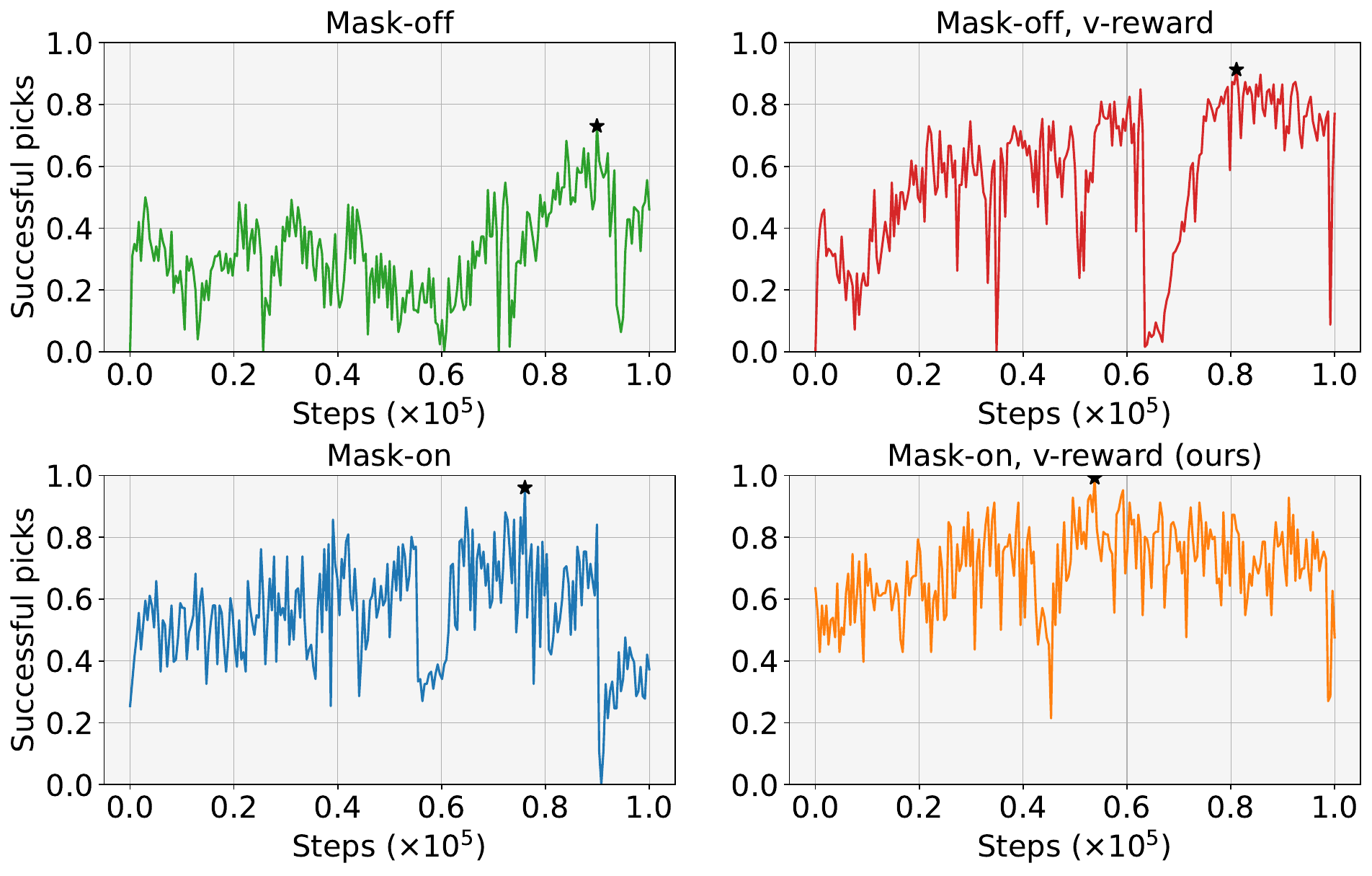}
        \caption{Training curves for the seeds with the best maximum performance.}
        \label{fig:best_max_app}
    \end{subfigure}
    ~
    \begin{subfigure}[t]{0.8\linewidth}
        \centering
        \includegraphics[width=0.7\textwidth]{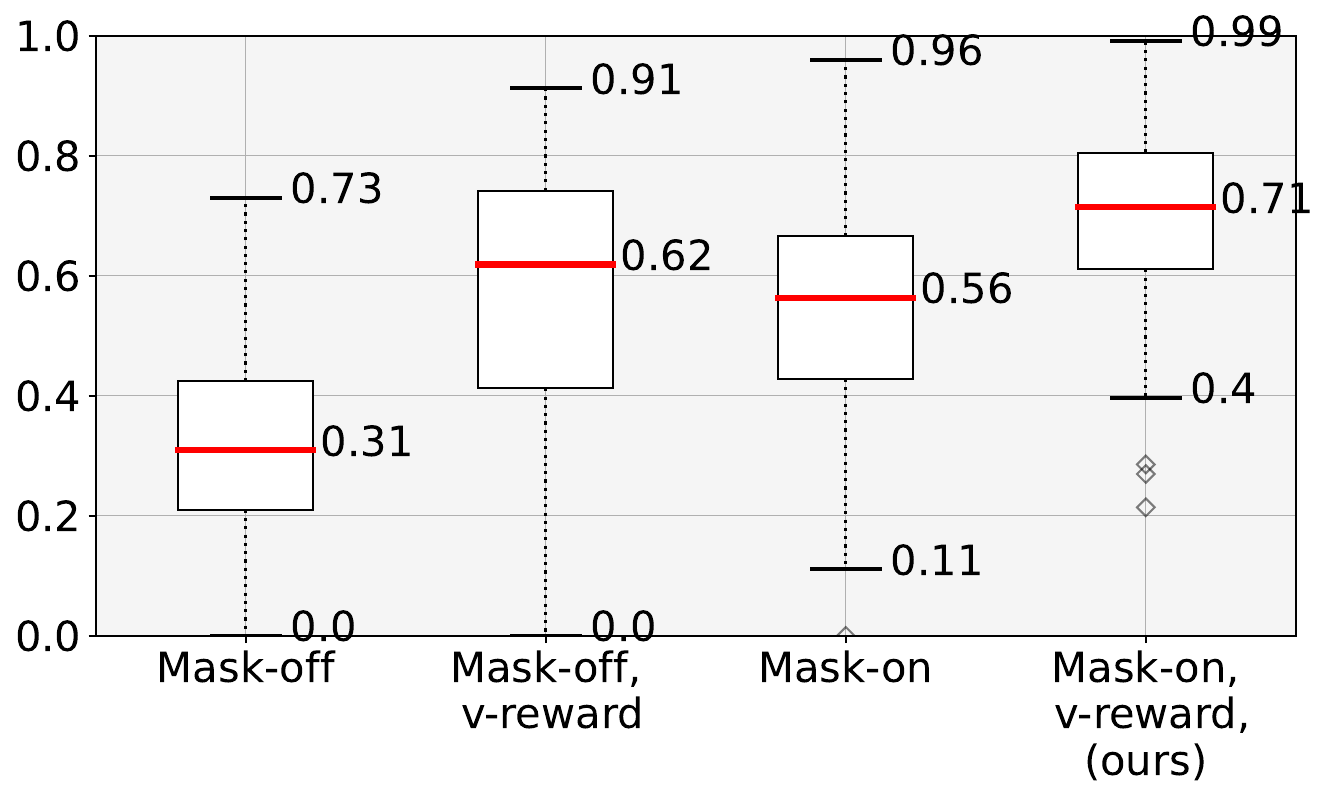}
        \caption{Box plot summarizing the results in Fig.~\ref{fig:best_max_app} where each evaluation step is treated as a data sample.}
        \label{fig:box_plot_best_max_app}
    \end{subfigure}
    \caption{Results for the seeds with the best maximum performance. For each version of our algorithm, the best seed is chosen among the seeds averaged in Fig.~\ref{fig:experiment_average}. The maximum performance is highlighted with "$\filledstar$". The experiments are conducted as in Fig.~\ref{fig:experiment_average} and the labels Mask-on and Mask-off denote the use of the safety bias layer in \eqref{eq:mask_func}, while v-reward denotes the use of $\mathcal{R}_v$ rather than $\mathcal{R}_b$ in \eqref{eq:reward_func}. These results are also illustrated in Fig.~\ref{fig:experiment_best} and \ref{fig:box_plot_best} and are enlarged here for additional clarity.}
    \label{fig:full_best_max_app}
\end{figure*}

\end{document}